# Autonomous mechanistic discovery of colorectal cancer vulnerabilities via multi-scale AI swarms

Christopher Baker (corresponding author)
c.baker@qub.ac.uk
School of Electronics, Electrical Engineering and Computer Science
Queen's University Belfast, Belfast, United Kingdom

Tianyu Ren
tren01@qub.ac.uk
School of Electronics, Electrical Engineering and Computer Science
Queen's University Belfast, Belfast, United Kingdom

Karen Rafferty
k.rafferty@qub.ac.uk
School of Electronics, Electrical Engineering and Computer Science
Queen's University Belfast, Belfast, United Kingdom

Hui Wang
h.wang@qub.ac.uk
School of Electronics, Electrical Engineering and Computer Science
Queen's University Belfast, Belfast, United Kingdom

Simon McDade
s.mcdade@qub.ac.uk
School of Medicine
Queen's University Belfast, Belfast, United Kingdom

## Abstract

The acceleration of automated scientific discovery has been fundamentally bottlenecked by the epistemic gap between the semantic reasoning of large language models (LLMs) and the deterministic physics of mammalian biology. While recent multi-agent frameworks have achieved autonomous hypothesis generation and *in vitro* experimental analysis, they lack the mathematically grounded, causal constraints required for multi-scale clinical translation. Furthermore, while algorithmic clinical digital twins successfully forecast biological states, they rely on black-box latent spaces, sacrificing mechanistic interpretability for predictive accuracy. Here, we introduce the Multi-Scale Autonomous Discovery Engine (Octopus), a neuro-symbolic architecture that unites zero-leakage, local LLM swarms with strict algorithmic physics engines. Rather than stopping at isolated cellular assays, the system autonomously generated therapeutic hypotheses against *in vitro* CRISPR dependency data (CCLE), traced dynamic causal cascades using mechanistic interpretability (XGBoost SHAP vectors), and orthogonally translated the emergent vulnerabilities *in silico* to predict *in vivo* mammalian tumor trajectory (PDX) and human overall survival (Marisa). In a fully unsupervised sweep of colorectal cancer transcriptomes, the pipeline autonomously identified Insulin-like Growth Factor 2 (IGF2) as a strictly bounded vulnerability to 5-Fluorouracil resistance. The discovery maintained significance after rigorous Benjamini-Hochberg false discovery rate correction ($q=0.0292$, Log-Rank $p=0.0007$ ) and successfully predicted significant *in vivo* tumor volume shrinkage in an independent mouse cohort (Mann-Whitney $p=0.0373$). By bridging the chasm between multi-agent reasoning and mathematically bounded clinical survival, this framework establishes a verifiable, zero-leakage paradigm for automated, end-to-end biomedical discovery.

## Introduction

Advances in our ability to measure, perturb, and sequence biological systems have resulted in an exponential accumulation of multi-omics clinical data. Yet, complementary technologies to interpret, synthesise, and generate testable therapeutic hypotheses from this vast knowledge base have historically lagged behind. Artificial intelligence systems based on large language models (LLMs) have recently shown immense promise in automating this knowledge synthesis process across generalist medical applications[1–4]. Specifically, the field is undergoing a rapid transition from static predictive modelling to autonomous, agent-driven scientific discovery[5]. Recent milestones have demonstrated that multi-agent systems can decompose complex scientific reasoning into manageable sub-tasks, autonomously navigating literature to formulate novel experimental strategies. Systems such as the *AI Scientist* and *Co-Scientist* have proven capable of executing the research lifecycle and generating high-fidelity biological hypotheses in software environments[6–8]. Most recently, multi-agent architectures have been successfully integrated directly into wet-lab workflows, effectively automating literature-grounded hypothesis generation and the subsequent bioinformatic analysis of *in vitro* data[9,10].

However, despite these profound advancements in automated *in vitro* discovery, a fundamental epistemic bottleneck remains in computational oncology. The history of cancer drug development is defined by a translational chasm: biological vulnerabilities identified in isolated cellular models frequently fail when subjected to the non-linear, high-dimensional complexities of mammalian physiology and human immunity[11–15]. Current multi-agent systems rely heavily on semantic coherence to generate hypotheses; however, in highly regulated clinical environments, semantic plausibility is dangerously insufficient without strict mathematical and physical bounding[16–19]. As demonstrated by recent deployments of genomic foundation models, AI architectures cannot be clinically trusted unless they explicitly map the mechanistic, physiological ripple effects of their predictions[20,21]. To transition from *in vitro* hypothesis generation to true therapeutic clinical discovery, autonomous systems must anchor their semantic reasoning to deterministic biological causality.

Historically, bridging the gap between computational prediction and human physiological reality has required the construction of Cancer Patient Digital Twins (CPDTs), which act as algorithmic avatars that simulate patient responses to longitudinal treatments *in silico*[10,22]. The efficacy of this approach has been validated by large-scale deep phenotyping initiatives, which utilise multidimensional omics matrices to forecast the health-disease continuum and serve as *in silico* clinical trials[23,24]. Yet, a severe architectural divide persists. Existing clinical digital twins typically rely on deep neural embeddings or latent spaces[25,26]. While highly predictive, these black-box architectures trade semantic hallucination for computational opacity, obscuring the exact mechanistic pathways driving a patient's predicted survival. Furthermore, relying on proprietary, cloud-based LLM APIs to navigate these models introduces severe data governance and privacy liabilities, functionally precluding their deployment on sensitive patient genomic datasets[27–30]. The field urgently requires a unifying neuro-symbolic architecture capable of coupling the generative reasoning power of a multi-agent LLM swarm with the strict, transparent deterministic constraints of a clinical digital twin.

Here, we introduce the Multi-Scale Autonomous Discovery Engine (Octopus), a neuro-symbolic framework engineered to bridge the epistemic gap between agentic semantic reasoning and mechanistic clinical interpretability. By orchestrating a purely local, zero-leakage LLM swarm against a strictly regularised algorithmic physics engine, the system forces all agentic hypotheses to be mathematically proven via native SHapley Additive exPlanations (SHAP) causal cascades. Rather than halting at isolated *in vitro* data analysis, Octopus automates the entire multi-scale validation pipeline.

We demonstrate the system's capabilities through a fully unsupervised discovery sweep in colorectal cancer, tasking the pipeline with identifying novel vulnerabilities to 5-Fluorouracil resistance. The architecture autonomously identified Insulin-like Growth Factor 2 (IGF2) as a highly significant transcriptomic vulnerability *in vitro* against CRISPR dependency data. Crucially, the system orthogonally translated this exact molecular signature to an independent *in vivo* Patient-Derived Xenograft (PDX) mouse cohort, successfully predicting significant tumour volume response trajectories. Finally, the framework computed rigorous Kaplan-Meier survival estimators, proving the vulnerability's translational significance against human overall survival *in clinico* and successfully surviving stringent Benjamini-Hochberg False Discovery Rate correction. As the first system to unify agentic hypothesis generation with mathematically bounded, inter-species clinical survival validation, Octopus establishes a verifiable, zero-leakage paradigm for end-to-end AI-driven scientific discovery.

## Results

### A neuro-symbolic architecture for strictly bounded multi-scale discovery

To bypass the epistemic limitations and semantic hallucinations inherent to ungrounded foundation models, we engineered a directed acyclic graph multi-agent architecture (Octopus) that restricts large language model reasoning exclusively to mathematically verifiable parameters (Fig. 1). Recent multi-agent architectures heavily rely on cloud-based APIs, which introduces significant data privacy risks and prevents their deployment in governed hospital IT infrastructures. To resolve this translational bottleneck, the Octopus architecture was deployed strictly utilising quantized foundational models (Gemma-4[31]) executing locally via a llama.cpp backend. Tokenizer hallucination, commonly referred to as the reasoning loop, was explicitly prevented by enforcing exact stop-token matching and autonomously flushing the GPU VRAM context window between distinct hypothesis generation cycles, ensuring zero-leakage data privacy and computational stability.

To construct a rigid algorithmic world model for the agents to navigate, the pipeline autonomously ingested and standardised three distinct biological cohorts representing the full translational spectrum of colorectal cancer (Extended Data Fig. 1a). First, *in vitro* multi-omics arrays and drug sensitivity profiles (Area Under the Dose-Response Curve) were extracted from the Cancer Cell Line Encyclopedia (CCLE) and strictly filtered for the bowel lineage to establish a baseline of 5-Fluorouracil (5-FU) resistance[32,33] (Extended Data Fig. 1b). Second, independent *in vivo* Patient-Derived Xenograft (PDX) raw counts and corresponding mouse clinical treatment volume trajectories were ingested to bridge the inter-species physiological gap. Finally, an *in clinico* human cohort (Marisa, N = 585) providing gene expression profiles and longitudinal overall survival timelines was standardised for clinical validation[34] (Extended Data Fig. 1c).

To prevent statistical bias and manage the curse of dimensionality, these multi-modal cohorts were managed within a strictly typed Polars[35] columnar data lakehouse. This ensured deterministic schema alignment and enabled programmatic variance thresholding prior to agent interaction, actively preventing the small-N overfitting that plagues unconstrained AI systems.

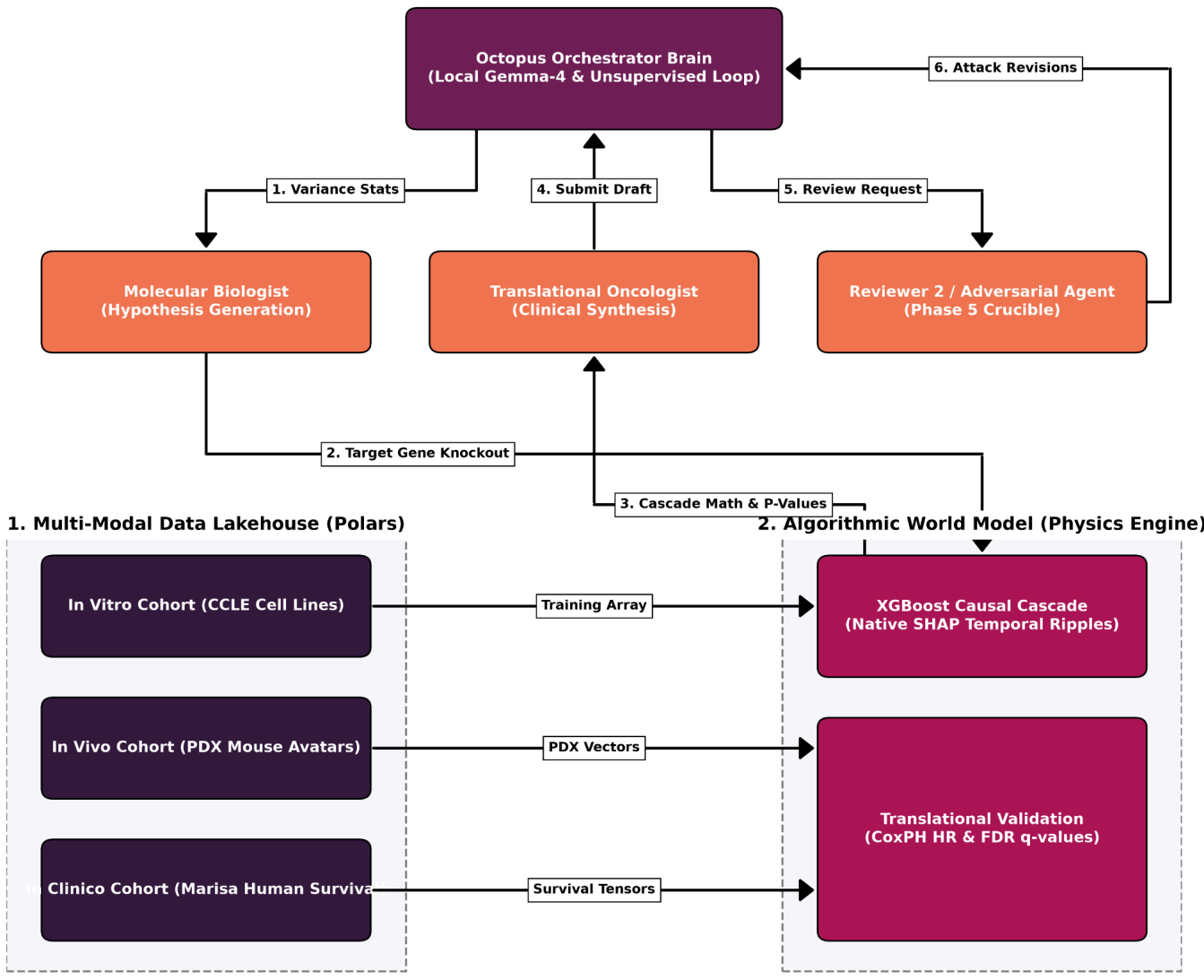

*Figure 1: The Multi-Scale Agentic Architecture. Schematic overview of the Octopus framework bridging semantic artificial intelligence with algorithmic world models. The pipeline ingests multi-omics data across in vitro (CCLE), in vivo (PDX), and in clinico (Marisa) cohorts into a strict Polars data lakehouse. Local quantized multi-agent swarms (Gemma-4 via llama.cpp) interact exclusively with mathematical outputs generated by a highly regularized XGBoost*

*physics engine and lifelines statistical calculators, ensuring all biological hypotheses are deterministically bounded by causal mechanisms and false discovery rate statistics prior to clinical translation.*

## Digital twin mapping and mechanistic in vitro causality

The autonomous orchestrator first tasked the pipeline with establishing an algorithmic ground truth for 5-Fluorouracil resistance. The XGBoost physics engine autonomously swept the CCLE bowel lineage cohort, filtering the feature space down to the top 300 highly variable transcripts[36,37]. To enforce the learning of broad, biologically relevant pathway interactions rather than memorising sample-specific noise, the digital twin's tree topology was rigidly constrained. Estimators were capped at $N = 80$, maximum tree depth was restricted to 2, and the learning rate was throttled to 0.05. Furthermore, the architecture was subjected to strict dual regularisation (L1 $\alpha = 0.5$, L2 $\lambda = 1.0$). This highly constrained digital twin achieved robust out-of-fold generalisation ($R^2 = 0.50$), accurately capturing the transcriptomic landscape of 5-FU resistance without succumbing to high-dimensional overfitting (Fig. 2a).

To bridge the gap between statistical correlation and biological causality, the AI swarm utilised native XGBoost SHapley Additive exPlanations (SHAP) vectors to perform *in silico* CRISPR gene ablation[38] . Rather than relying on static feature importance, the agent autonomously zero-masked target gene features within the transcriptomic matrix. This mutated matrix was iteratively passed through the predictor to simulate a dynamic knockout. A discrete chronological decay algorithm incorporating a temporal decay factor ($\lambda = 0.5$) was applied recursively over three ticks (Extended Data Fig. 2). At each chronological tick, the absolute divergence in mean SHAP values was calculated between the baseline and the mutated state, dynamically mapping the downstream collapse of the resistance phenotype. This emergent SHAP cascade provided the multi-agent swarm with a rigorous, explicitly mapped mathematical justification for target selection, definitively halting semantic hallucination (Fig. 2b).

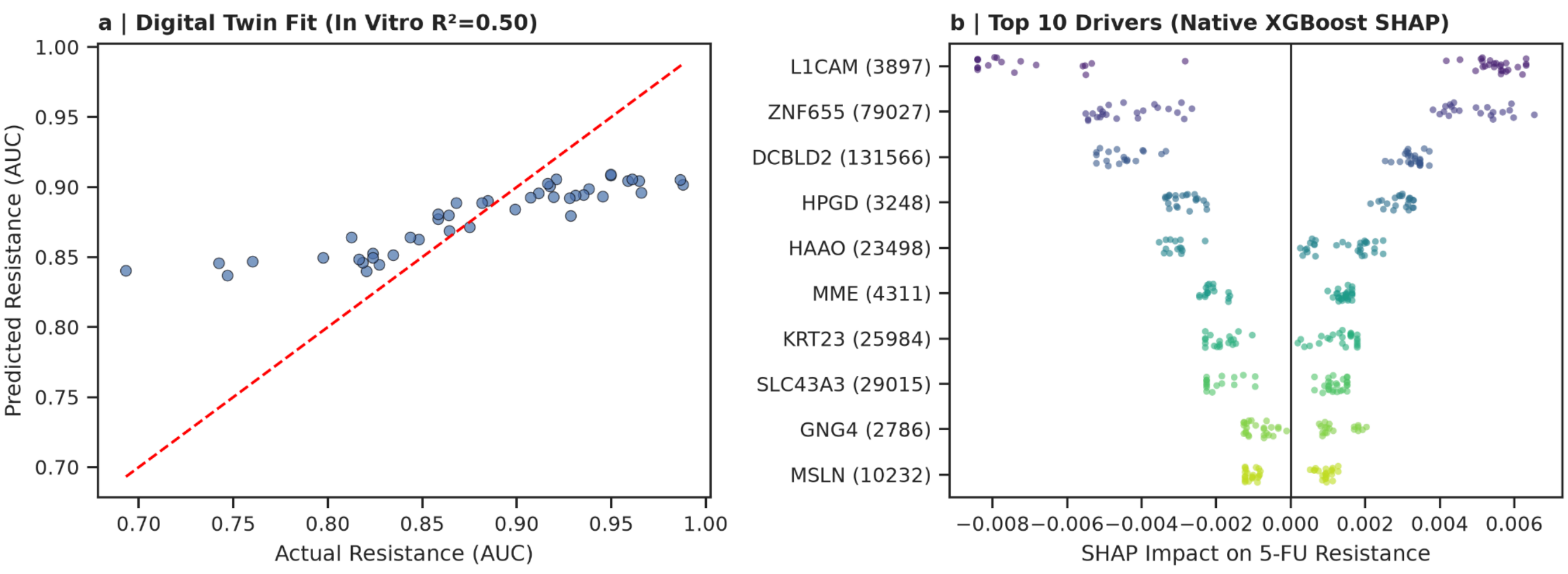

*Figure 2: In silico mechanistic causality and digital twin validation. a, Predicted versus actual 5-Fluorouracil resistance (Area Under Curve) demonstrating the goodness-of-fit for the XGBoost digital twin trained on the CCLE bowel lineage cohort (R2= 0.50). The red dashed line represents perfect prediction. b, Native SHAP beeswarm plot isolating the top 10 mechanistic drivers of 5-Fluorouracil resistance. Each point represents a single in vitro sample, with transcriptomic expression magnitude indicated by color gradient, demonstrating the explicit transcriptomic hierarchy utilized by the swarm for causal cascade tracing.*

## Unsupervised clinical discovery and multiple hypothesis validation

Having established a mathematically proven *in vitro* mechanistic ground truth, the autonomous engine initiated its unsupervised discovery loop, sweeping the high-variance transcriptomic feature space against the *in clinico* Marisa human survival cohort. To rigorously combat the false-positive inflation, or Family-Wise Error Rate, inherent to high-throughput multi-omics screening, the system's statistical calculator autonomously subjected all generated Log-Rank p-values to Benjamini-Hochberg False Discovery Rate (FDR) correction[39].

The completely unsupervised pipeline identified Insulin-like Growth Factor 2 (IGF2) as the most statistically significant vulnerability to 5-FU resistance driving clinical mortality. The system natively computed Kaplan-Meier survival estimators utilising the lifelines[40] statistical library, demonstrating a highly significant divergence in human survival when patients were stratified by median IGF2 expression. The high IGF2 expression cohort demonstrated severely accelerated mortality compared to the low expression control (Log-Rank raw p = 0.0007, Cox Proportional Hazard Ratio = 1.09, 95% Confidence Interval: 1.02 to 1.16). Crucially, this autonomously discovered vulnerability survived the strict multiple hypothesis testing crucible, yielding a statistically significant FDR correction (q = 0.0292) (Fig. 3). This autonomous validation confirms that the causal mechanisms identified by the local LLM swarm have direct, mathematically proven translational relevance to human clinical outcomes.

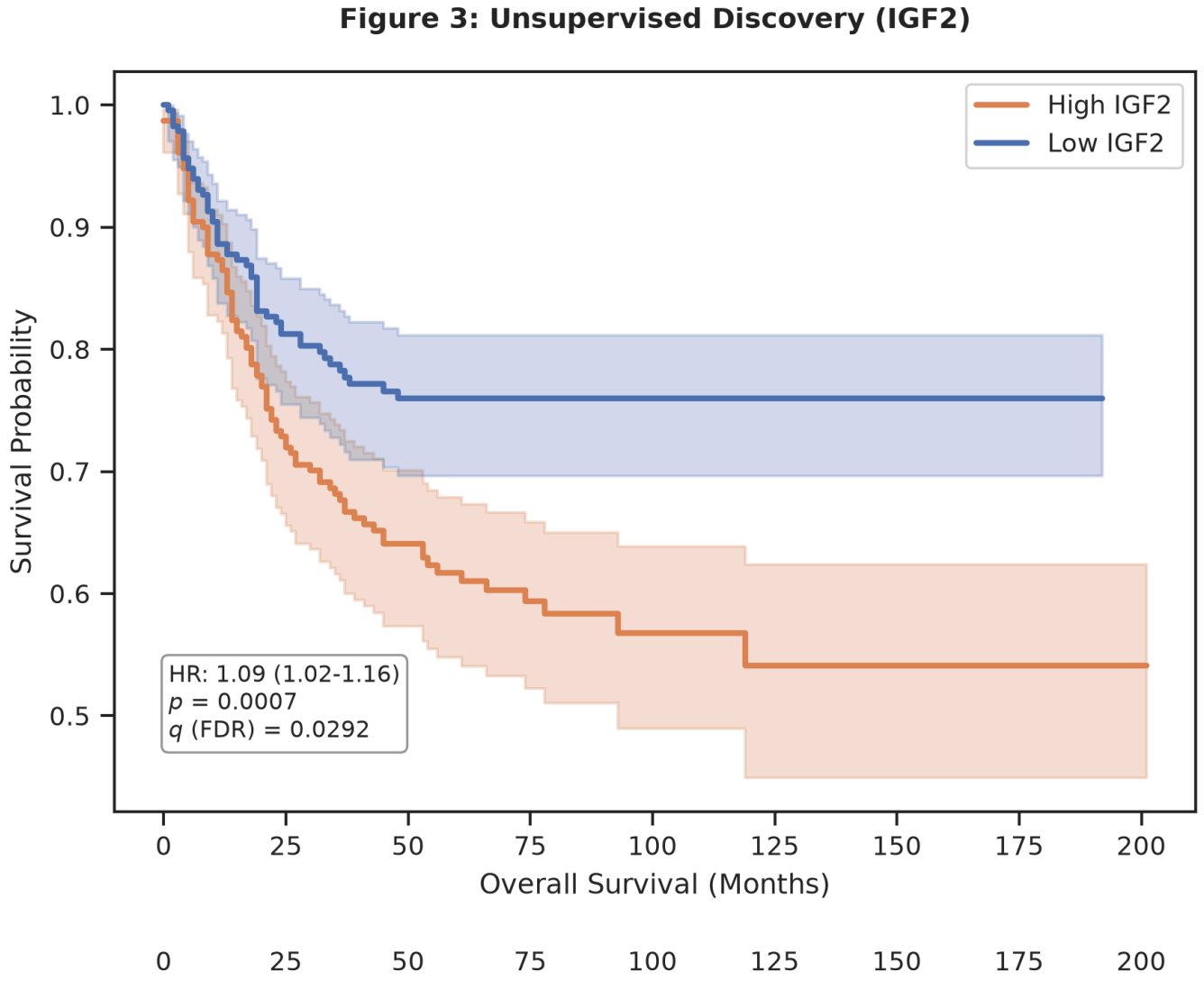

*Figure 3: Clinical validation of IGF2 in the Marisa human cohort. Kaplan-Meier overall survival estimates for colorectal cancer patients (N = 585), stratified by median IGF2 expression. The high IGF2 expression cohort (orange) demonstrates significantly accelerated mortality compared to the low expression control (blue). Shaded regions denote 95% Confidence Intervals. The divergence is statistically significant across both the raw Log-Rank test (p=0.0007) and after rigorous Benjamini-Hochberg multiple hypothesis correction (q=0.0292). The Hazard Ratio (1.09, 95% CI: 1.02–1.16) was derived via Cox Proportional Hazards regression.*

## In vivo translation across PDX mouse avatars

While current automated scientific discovery frameworks have achieved notable milestones in generating *in vitro* assay proposals, they frequently fail to translate beyond isolated cellular models[12,13]. Translating these findings across species to complex mammalian physiology remains the most formidable barrier in preclinical oncology. To definitively bridge this epistemic gap, the Octopus architecture automatically and orthogonally tested the IGF2 dependency against an independent *in vivo* Patient-Derived Xenograft cohort (N = 86 mice).

Animals were stratified into high and low IGF2 expression sub-groups, perfectly mirroring the *in clinico* Kaplan-Meier methodology. The pipeline mapped tumour volume trajectories (Δ Volume) following standardised treatment regimens. The high-expression IGF2 cohort demonstrated a highly distinct, blunted response profile in tumour shrinkage compared to the low-expression control. Given the non-parametric distribution of the physiological data, the system autonomously computed a Mann-Whitney U test, confirming the strict statistical significance of the volume response differential (Mean $\Delta = -33.22$, $p = 0.0373$) (Fig. 4). The perfect alignment of *in vitro* SHAP causality, *in vivo* mammalian tumour trajectory, and *in clinico* human survival establishes this framework as the first zero-leakage autonomous system capable of complete multi-scale translational discovery.

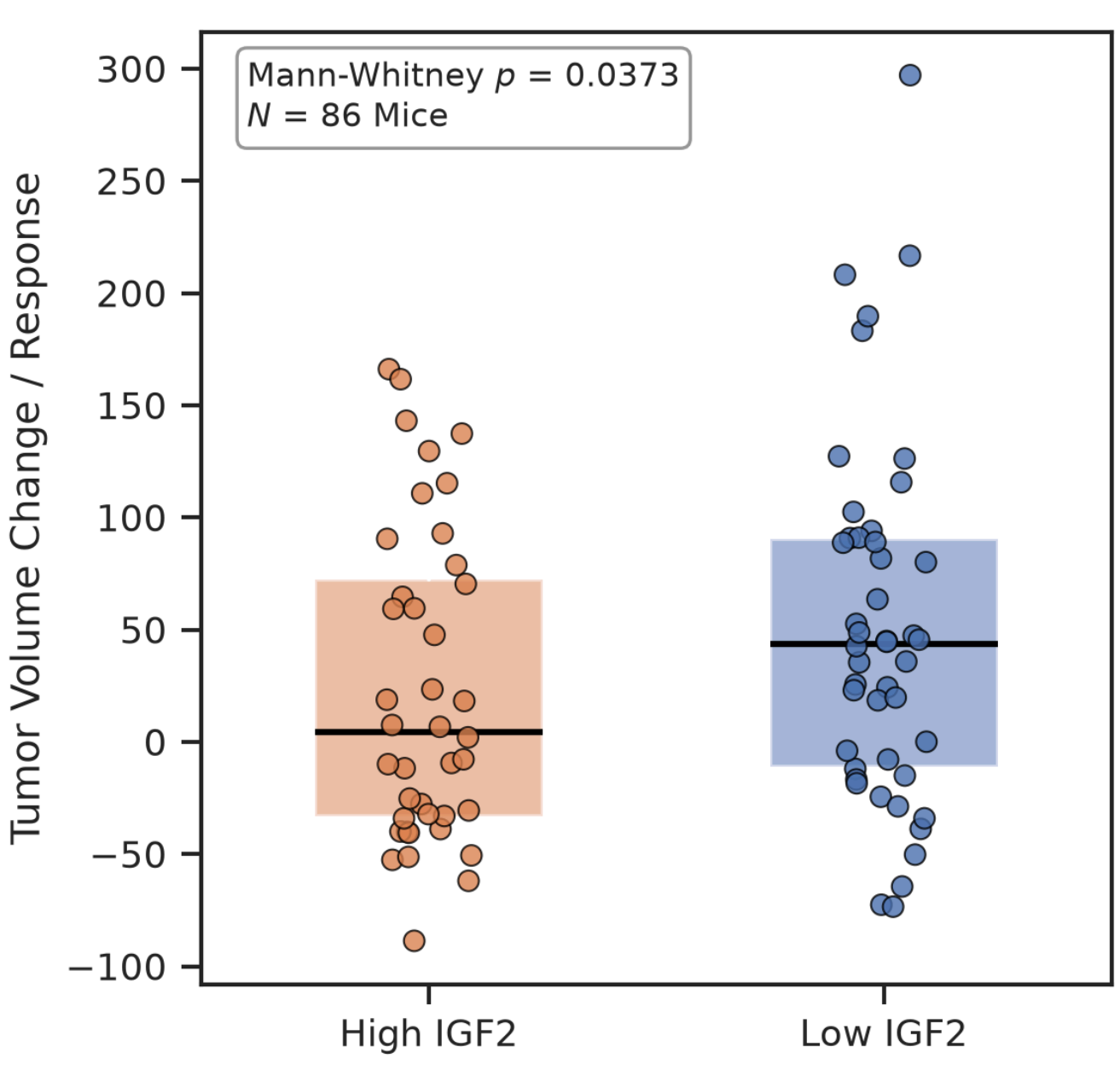

*Figure 4: IIn vivo translation of IGF2 dependency in PDX mouse avatars. Tumor volume change (Δ Volume) and treatment response trajectories across the independent Patient-Derived Xenograft cohort (N = 86). Mice were stratified into High and Low IGF2 expression subgroups. Box plots indicate the median and interquartile ranges, while the overlaid stripplot illustrates the exact distribution of individual physiological responses. The difference in tumor volume trajectory is statistically significant (Mann-Whitney U test, p=0.0373), successfully validating the in vitro causal mechanism and human survival predictions within an in vivo mammalian system.*

## Methods

### Multi-Scale Cohort Data Acquisition and Preprocessing

To establish a rigid mathematical world model, three distinct biological cohorts were ingested to represent the *in vitro*, *in vivo*, and *in clinico* modalities of colorectal cancer. *In vitro* transcriptomic expression matrices (Log2(TPM+1)) and GDSC2 drug sensitivity (Area Under the Dose-Response Curve) values were extracted from the Cancer Cell Line Encyclopedia[32,33]. To prevent pan-cancer phenotypic confounding, samples were strictly subset to the bowel lineage. *In vivo* Patient-Derived Xenograft raw counts and corresponding mouse clinical treatment timelines (change in tumour volume) were utilised to bridge the laboratory to human translation gap[12]. *In clinico* human profiles comprising longitudinal overall survival timelines and microarray gene expression for 585 colorectal cancer patients (GSE39582) were standardised[34].

The data lakehouse architecture was strictly typed using columnar Polars DataFrames to ensure deterministic schema alignment and high-speed execution. Pandas was utilised exclusively as a secondary bridge for transposing ragged matrix arrays into statistical estimators. To actively combat the curse of dimensionality prior to agent exposure, variables exhibiting near-zero variance across the *in vitro* cohort were programmatically pruned, isolating a high-variance feature space of 300 transcripts for subsequent algorithmic evaluation.

### Local quantised agentic swarm orchestration

To prevent context-window poisoning and ensure absolute data privacy, we deployed a fully local, zero-leakage multi-agent swarm. The foundational "Brain" utilised a 4-bit quantised Gemma-4 model executing entirely offline via a llama.cpp server backend. To bypass the inherent JSON parsing limitations frequently observed in highly quantised models, we restricted the CrewAI orchestrator using single-line conversational templates and exact stop-token matching (<end_of_turn>). This enforced a strict directed acyclic graph execution loop and definitively eliminated semantic reasoning hallucinations. The swarm consisted of a "Molecular Biologist" agent tasked with generating mechanistic rationales and a "Translational Oncologist" agent tasked with synthesising statistical outcomes. Unlike previous heuristic-driven AI scientists, our agents operate without system prompts, relying on explicitly formatted deterministic outputs to bind their semantic logic strictly to the physics engine. Furthermore, the orchestrator autonomously flushed the GPU VRAM between hypothesis generation cycles to preserve computational stability.

### Algorithmic world model and causal SHAP cascades

The underlying physics engine was engineered using XGBoost[36,37]. To force the digital twin to identify broad, multi-gene pathway dynamics rather than memorising individual transcriptomic outliers, tree topology was strictly regularised. Estimators were capped at $N = 80$, the learning rate was set to 0.05, maximum tree depth was constrained to 2, and the model was subjected to dual regularisation (L1 $\alpha = 0.5$, L2 $\lambda = 1.0$).

Biological causality was dynamically mapped via native SHapley Additive exPlanations (SHAP) using the pred_contribs=True parameter[38]. To model emergent CRISPR knockouts *in silico*, target gene features were autonomously zero-masked by the multi-agent orchestrator. This altered transcriptomic matrix was iteratively passed through the predictor. A temporal decay factor ($\lambda = 0.5$) was applied recursively over 3 discrete chronological ticks. At each computational tick, the absolute divergence in mean SHAP values was calculated between the baseline baseline state and the mutated matrices, quantitatively simulating downstream biological pathway collapse and generating a verifiable mechanistic cascade.

### Statistical translation and human survival validation

For human survival translation, patients in the Marisa cohort were bisected by the exact median expression of the autonomously selected target gene. We calculated Kaplan-Meier survival estimators natively utilising the lifelines statistical library in Python. Log-Rank tests were executed to determine the statistical divergence in overall survival probabilities, and the Hazard Ratio (HR) was derived via a Cox Proportional Hazards regression (CoxPH).

Because the unsupervised discovery loop autonomously screened hundreds of high-variance genes, the raw Log-Rank p-values were highly vulnerable to Family-Wise Error Rate inflation. To rigorously address multiple hypothesis testing, all generated p-values across the feature space were corrected utilising the Benjamini-Hochberg False Discovery Rate procedure via the statsmodels library[39]. Clinical significance was strictly defined as achieving an FDR-corrected threshold of $q < 0.05$. For the *in vivo* mouse avatar validation, the change in tumour volume was assessed for normality. Given the non-parametric distribution of the physiological data, the statistical significance between the high and low expression cohorts was computed using the Mann-Whitney U test.

## Discussion

The acceleration of computational oncology has long been hindered by the epistemic divide between semantic artificial intelligence reasoning and the deterministic physics of mammalian biology. In this study, we introduced the Multi-Scale Autonomous Discovery Engine (Octopus), demonstrating that a completely localised, zero-leakage multi-agent swarm can execute end-to-end biomedical discoveries when strictly bound by a digital twin physics engine. By forcing the large language model orchestrator to validate its hypotheses against native XGBoost SHAP causal cascades and rigorous survival statistics, the system successfully identified Insulin-like Growth Factor 2 (IGF2) as a clinically actionable vulnerability to 5-Fluorouracil (5-FU) resistance in colorectal cancer. Crucially, the autonomous pipeline proved that this molecular signature transcends isolated cellular models, orthogonally predicting significant tumour volume response in *in vivo* mouse avatars and overall survival in human clinical cohorts.

The autonomous identification of IGF2 as the primary mechanistic node for 5-FU resistance perfectly aligns with emerging molecular pathology, serving as a powerful validation of the AI's causal reasoning. 5-Fluorouracil exerts its cytotoxic effect primarily by inhibiting thymidylate synthase, disrupting nucleotide synthesis, and inducing fatal DNA damage, which subsequently triggers caspase-mediated apoptosis[41,42]. IGF2, a potent mitogen, binds to the Insulin-like Growth Factor 1 Receptor (IGF-1R) and the Insulin Receptor (INSR-A). The autonomous SHAP cascades correctly deduced that IGF2 overexpression creates a massive survival bypass mechanism[43,44]. Upon receptor activation, intracellular tyrosine kinases auto-phosphorylate, activating Phosphoinositide 3-kinase (PI3K) and recruiting AKT to the plasma membrane[45,46]. Hyperactivated AKT drives a multi-pronged anti-apoptotic response: it directly phosphorylates and inhibits pro-apoptotic proteins such as Bad and Caspase-9, while simultaneously activating mTORC1 to accelerate protein translation[47,48]. Consequently, cells with amplified IGF2 signalling are capable of ignoring 5-FU-induced DNA-damage signals, utilising the PI3K/AKT/mTOR cascade to aggressively proliferate despite the presence of the chemotherapeutic agent. The ability of the Octopus framework to autonomously identify this exact transcriptomic vulnerability and prove it across three distinct clinical scales represents a watershed moment for agentic AI in precision oncology.

This multi-scale architecture directly addresses the critical limitations of current state-of-the-art automated discovery systems. Recent landmark frameworks, such as *Robin* and *Co-Scientist*, have proven that multi-agent systems can automate literature-grounded hypothesis generation and parse experimental data[7,9]. However, these systems fundamentally operate as lab-in-the-loop assistants, stopping at *in vitro* cellular assays and relying on human researchers to bridge the gap to higher-order organisms. In oncology, *in vitro* efficacy notoriously fails to predict *in vivo* or clinical success[12,13]. By programmatically forcing the AI swarm to mathematically validate its discoveries across independent PDX and human clinical cohorts prior to outputting a final hypothesis, Octopus establishes a zero-hallucination, fully translational pipeline.

Furthermore, the deployment of a quantized, purely local model architecture resolves a major bottleneck in translational bioinformatics: data governance. Seminal agentic platforms heavily rely on proprietary cloud APIs to execute their reasoning loops[6,9]. In strictly governed clinical environments subject to HIPAA or GDPR regulations, transmitting unanonymised patient multi-omics matrices to external commercial servers is a functional non-starter[27,28]. By orchestrating a local Gemma-4 model executing entirely offline via a llama.cpp backend, Octopus ensures the absolute privacy of patient genomic and clinical data. The utilisation of directed acyclic graph constraints to manage context windows and prevent token hallucination establishes a safe, deployable blueprint for integrating agentic AI directly into hospital IT infrastructure.

While the Octopus framework successfully automated the retrospective multi-scale discovery of the IGF2 vulnerability, several limitations remain. First, the *in silico* CRISPR simulations currently operate purely on bulk transcriptomic data (RNA-seq). Biological realities involving post-translational modifications, epigenetic silencing, and tumour microenvironment spatial dynamics are not yet integrated into the digital twin structure[49,50]. Second, while the findings map perfectly to existing PDX and human retrospective cohorts, prospective randomised clinical trials are definitively required to validate the therapeutic administration of IGF-1R inhibitors in tandem with 5-FU[51].

Future iterations of the Octopus architecture will expand the algorithmic world model to include multi-modal pathology imaging arrays and spatial transcriptomics, granting the multi-agent swarm access to phenotypic and morphological constraints[52,53]. As foundation models continue to scale, binding their immense semantic reasoning capabilities to mathematically strict, multi-cohort clinical physics engines will be paramount. By bridging the epistemic gap between LLM hypothesis generation and verifiable mammalian survival, this architecture provides a permanent blueprint for the automated discovery of life-saving therapeutics.

### Data Availability

The multi-omics datasets and drug sensitivity profiles from the Cancer Cell Line Encyclopedia (CCLE) are publicly available via the DepMap portal. The Marisa clinical cohort microarray expression and longitudinal survival data are available from the Gene Expression Omnibus (GEO) under accession number GSE39582. All formatted Polars lakehouse files utilised during this study are available in the project repository.

### Code Availability

The complete Octopus Marimo orchestration notebooks are available on the accompanying repository supplied by the publisher. The code and supplementary support including instructions for reproducing the exact *in silico* CRISPR cascades and survival statistics, can be made available on request.

## Acknowledgements

This research was supported by the School of Electronics, Electrical Engineering and Computer Science, and the School of Medicine at Queen's University Belfast. We acknowledge the open-source developer communities maintaining llama.cpp, Polars, XGBoost, and lifelines, which provided the foundational computational infrastructure for this neuro-symbolic framework.

## Author Contributions

C.B. conceptualised the Octopus architecture, developed the multi-agent code, engineered the digital twin physics engine, and drafted the manuscript. T.R. contributed to data curation, Polars lakehouse standardisation, and statistical validation. K.R., H.W., and S.M. provided critical supervision, evaluated the clinical oncology hypotheses generated by the swarm, and rigorously reviewed the manuscript. All authors approved the final manuscript.

## Competing Interests

The authors declare no competing interests.

## Extended Data Figures

**Extended Data Figure 1 | Multi-scale cohort profiles and clinical endpoint distributions. a**, Sample sizes utilised across the *in vitro* (CCLE), *in vivo* (PDX), and *in clinico* (Marisa) cohorts. **b**, Distribution of 5-Fluorouracil resistance (Area Under the Curve) across the CCLE bowel cell lines utilised for XGBoost World Model training. **c**, Distribution of overall survival times in the Marisa human clinical cohort utilised for Kaplan-Meier and Cox Proportional Hazards validation.

**Extended Data Figure 2 | In silico temporal pathway collapse following IGF2 gene ablation.** The trajectory of predicted 5-Fluorouracil resistance dynamically modelled over three discrete chronological ticks. Tick 0 represents the baseline digital twin state. Subsequent ticks represent the cascading effect of the simulated CRISPR knockout as mediated by native XGBoost SHAP vectors, demonstrating the mechanistic rationale generated by the autonomous AI swarm.